\pdfoutput=1
\documentclass[11pt,a4paper]{article}

\PassOptionsToPackage{svgnames}{xcolor}
\usepackage{style}

\usepackage[numbers,sort&compress]{natbib}
\usepackage{graphicx}
\usepackage[font=small,labelfont=bf]{caption}
\usepackage{subcaption}
\usepackage{array}
\usepackage{booktabs}
\usepackage{colortbl}
\usepackage{multirow}
\usepackage{bigstrut}
\usepackage{float}
\usepackage[all]{hypcap}
\usepackage{pifont}
\usepackage{amsmath}
\usepackage{mathtools}
\usepackage{mathrsfs}
\usepackage{nicefrac}

\usepackage{algorithm}
\usepackage{algpseudocode}

\usepackage{microtype}
\usepackage{enumitem}
\usepackage{cleveref}
\usepackage{bxcoloremoji}
\usepackage{url}
\usepackage{wrapfig}
\usepackage{lipsum}
\usepackage{stackengine}
\usepackage{adjustbox}
\usepackage{rotating}
\usepackage{makecell}

\newcommand{\caer}{\textsc{CAER}}
\newcommand{\emptyact}{\emptyset}

\begin{document}

\Cover{
  title={CAER: Causal Action Effect Reweighting for World Model Training},
  authors={Jianjie Fang\textsuperscript{1,*}, Xvyuan Liu\textsuperscript{1,*}, Ziyou Wang\textsuperscript{1}, Rongze Tang\textsuperscript{3}, Zhaolu Wang\textsuperscript{1}, Zhuohang Li\textsuperscript{1}, Xin Zhang\textsuperscript{2}, Haisheng Su\textsuperscript{2}, Chen Gao\textsuperscript{1,\textdagger}, Wei Wu\textsuperscript{2}, Xinlei Chen\textsuperscript{1,\textdagger}, Yong Li\textsuperscript{1,\textdagger}},
  affils={\textsuperscript{1}Tsinghua University \quad \textsuperscript{2}Manifold AI \quad \textsuperscript{3}University of Science and Technology of China\\
  \texttt{chgao96@gmail.com},\quad \texttt{chen.xinlei@sz.tsinghua.edu.cn},\quad \texttt{liyong07@tsinghua.edu.cn}\\
  \textbf{\textsuperscript{*}Equal contribution. \quad \textsuperscript{\textdagger}Corresponding authors.}},
  abstract = {\textbf{Abstract.} World models are becoming core infrastructure for embodied intelligence, with action-conditioned video generation providing controllable predictions of how scenes evolve after agent interventions. Yet existing models are commonly trained with space--time-uniform mean squared error, allowing abundant background tokens to dominate the gradient while sparse interaction dynamics remain under-optimized; such uniform fitting rewards reconstructing appearance rather than learning how actions change the world. We introduce \textbf{C}ausal \textbf{A}ction \textbf{E}ffect \textbf{R}eweighting \textbf{(\caer{})}, a general training paradigm that redistributes supervision toward the tokens whose predicted future is causally affected by the action. \caer{} contrasts the model's own predictions with and without action conditioning to localize these tokens online, then normalizes the resulting effect map into a weight that preserves the total coefficient mass and changes only where it is spent. This online signal requires no external annotations or offline preprocessing, avoids additional data-processing time, and scales naturally with model and dataset size. Experiments across heterogeneous action-conditioned world-model tasks show that \caer{} converges to better solutions than uniform MSE training, with consistent improvements in the physical consistency, controllability, and visual quality of generated videos.},
  date={August 31, 2026},
  codeurl={https://github.com/EmbodiedCity/CAER.code},
  pageurl={https://manifoldai-research.github.io/CAER/},
  headerlogo={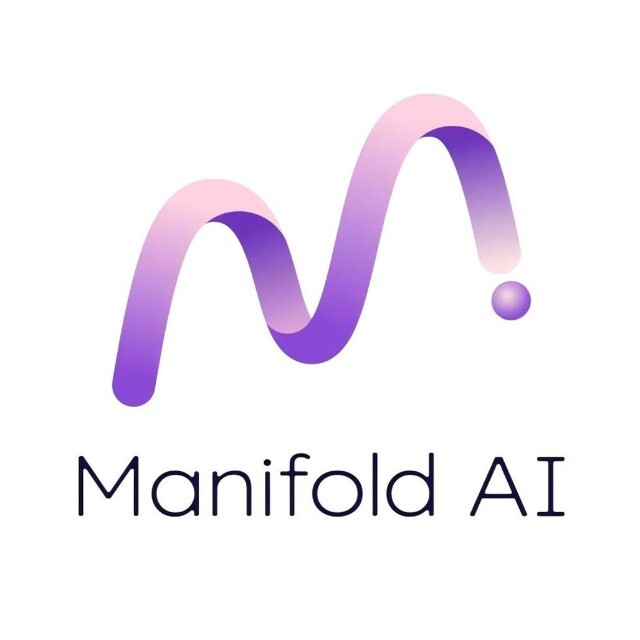},
  headerdate={August 31, 2026}
}

\section{Introduction}
\label{Intro}
World models equip embodied agents with the ability to look beyond the current observation and anticipate how the environment may evolve under different behaviors. By turning learned dynamics into simulated experience, they support planning, evaluation, and policy learning before actions are executed in the physical world. Advances in video diffusion and flow-matching backbones~\cite{wan,hunyuanvideo,cosmos} have made visually realistic world simulation increasingly feasible, supporting robotics, interactive environments, and policy development beyond what can be safely or economically collected in the physical world~\cite{cosmospredict,team2025gigaworld,fang2026worldscapemoe,zhao2026worldvln,su2026worldscapepolicy,fang2026iworldbench}. Within this paradigm, action-conditioned video generation provides a direct route to controllable prediction by modeling how visual scenes evolve in response to agent interventions. For an embodied agent, however, plausible appearance alone is insufficient. A useful world model must capture the physical consequences of interaction: where contact occurs, when an object begins to move, how forces propagate through the scene, and how the environment responds over time.

\begin{figure}[t]
\centering
\includegraphics[width=\linewidth]{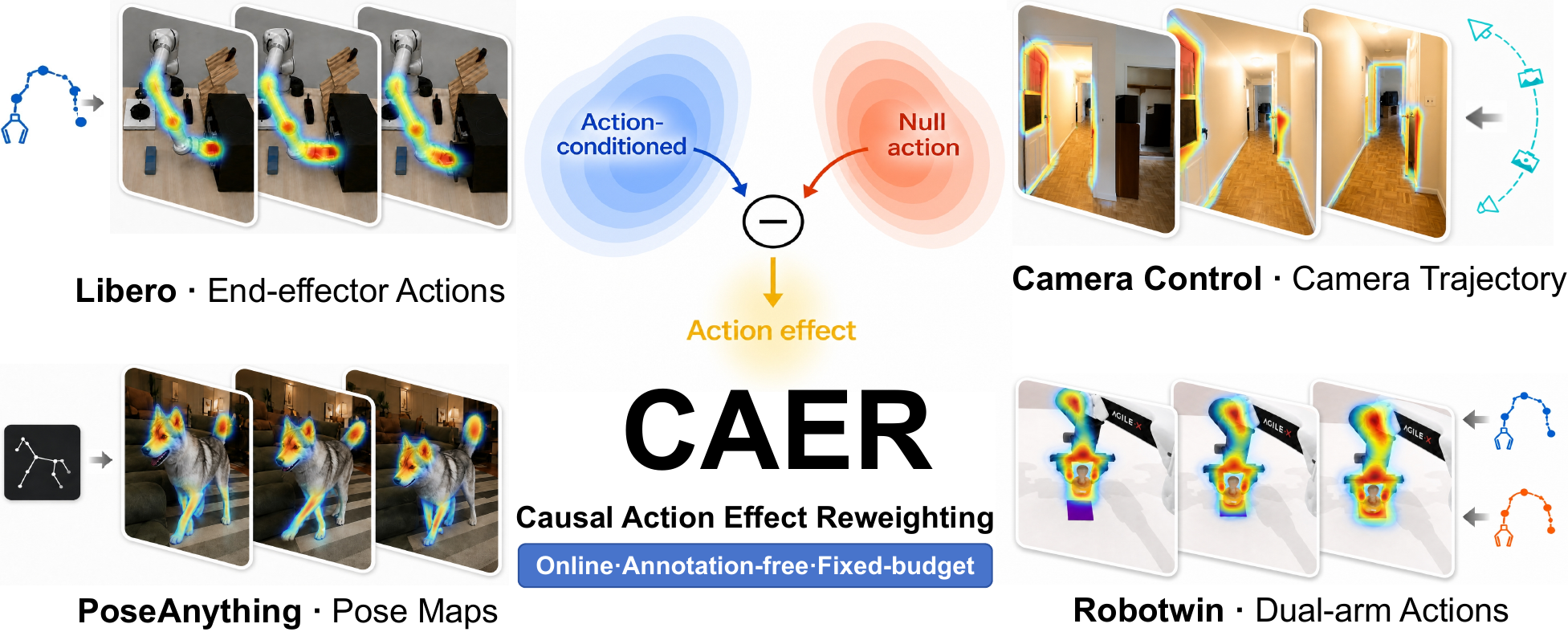}
\caption{\textbf{\caer{} overview.} \caer{} identifies and reweights action-responsive tokens online without external annotations. Examples span end-effector trajectories, camera trajectories, dual-arm actions, and pose maps.}
\label{fig:teaser}
\end{figure}

Despite this progress, current world models are still predominantly optimized with mean squared error averaged uniformly over space and time. Most video tokens describe static backgrounds, slowly varying context, or regions unrelated to the current interaction, whereas the decisive prediction errors are concentrated around brief contact events, manipulated objects, and actor motion. Uniform averaging therefore allows abundant, easily reconstructed tokens to dominate the gradient and dilute supervision for sparse physical interactions. This mismatch is especially damaging in embodied settings: a rollout may look realistic on average while placing contact at the wrong time, moving an object in the wrong direction, or failing to respond to the commanded action. Such failures are not merely visual defects; they indicate that the model has not learned how actions causally alter the world.

One possible remedy is to annotate actors, objects, motion, and contact regions explicitly. Segmentation models such as SAM~\cite{kirillov2023sam} and optical-flow estimators such as RAFT~\cite{teed2020raft} can track these regions and provide handcrafted loss weights. This solution, however, is difficult to scale to world-model data: per-frame inference introduces substantial computation, storage, and preprocessing overhead, while the resulting supervision inherits the biases and failure modes of the external models. It also defines relevance through a fixed annotation pipeline rather than through the causal relationship between the action and the predicted future.

In this work, we ask whether the world model itself can provide the missing supervision signal. Our key observation is that additional gradient should be assigned not to every visually salient region, but to tokens whose predicted future is sensitive to the action. If removing the action changes the prediction at a token, that token captures an action-dependent consequence that the model must learn, whereas a token that is unaffected by the action carries no action-conditional risk however much it moves. This turns loss allocation into an online causal action-response problem. The model needs no external definition of the actor, manipulated object, or contact interface, because it identifies relevant interactions through its own conditional response. This principle is action-agnostic: it applies to end-effector poses, joint commands, spatial action maps, and camera trajectories alike, whenever a null-action condition can be learned, which aligns it with the broader goal of grounding world models in the consequences of intervention rather than only in perceptual likelihood~\cite{lecun2022path,pearl2009causality}.

To examine whether self-computed causal action effects can serve as a scalable reweighting signal for world model training, we investigate three research questions.
\begin{itemize}
\item \textbf{RQ1: Does action-effect reweighting improve interaction-region prediction quality?}
Since token-uniform losses are the default for generative world models, it remains unclear whether reweighting toward unresolved action effects improves the regions that matter most---contact events, manipulated objects, and actor motion---while preserving background fidelity and overall visual quality. Answering this question establishes whether \caer{} is an effective replacement for uniform reconstruction loss in action-conditioned settings.
\item \textbf{RQ2: How do the key design choices contribute to the effect map quality and training stability?}
\caer{} introduces two training choices that govern the quality of the online counterfactual signal: the action-dropout rate that defines the null branch, and the fixed noise level at which the effect is read. We vary each in isolation to understand its contribution to convergence speed, loss stability, and final generation performance.
\item \textbf{RQ3: Does the self-computed signal sharpen as the model improves?}
A central motivation of self-computed reweighting is that the supervision signal improves together with the model, forming a training loop that needs no external annotation. We test whether the benefit of the self-computed signal grows with training, overtaking uniform averaging once the causal interaction is learned, and whether the resulting gains appear in task-level success rather than in appearance metrics alone.
\end{itemize}

\begin{figure}[t]
\centering
\includegraphics[width=\linewidth]{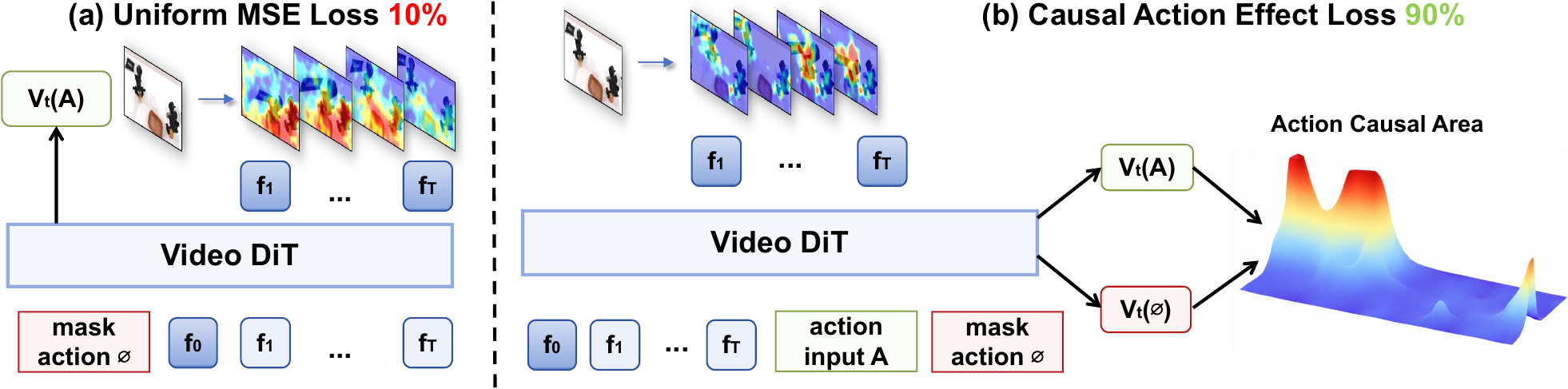}
\caption{\caer{} framework overview. At a fixed intermediate noise level and under shared noise, \caer{} contrasts the model's response under the real action and a learned null action to localize action-sensitive tokens, then normalizes the map sample-wise to unit mean. The objective redistributes a fixed coefficient mass toward causally responsive regions, while action dropout keeps every token stochastically supervised.}
\label{fig:caer_framework}
\end{figure}

Building on this observation, we introduce \textbf{\caer{}} (Fig.~\ref{fig:teaser}), a causal action effect reweighting paradigm that changes how the conventional MSE-based flow-matching objective allocates supervision. Instead of averaging all space--time tokens uniformly, \caer{} compares predictions under the observed action and a learned null-action condition at a fixed intermediate noise level, using their velocity-space difference to identify where the predicted future responds to the action. Action dropout keeps this counterfactual query in distribution. The action response is then normalized sample-wise into a weight that preserves the total coefficient mass and changes only where it is spent. Because the coefficient mass is fixed, the comparison with uniform averaging becomes a pure allocation question: to first order in the learning rate, focused weighting reduces interaction-region risk faster than uniform averaging whenever the weight is positively correlated with token utility, and it still descends at points where uniform training stalls because sparse interaction gradients are cancelled by dominant background gradients. Because the complete signal is computed online by the world model, \caer{} requires no segmentation, tracking, optical flow, or contact annotation and applies to both spatial and numerical action interfaces.

Experiments show that this causal redistribution improves interaction modeling and visual generation together. Because the weight is normalized rather than added, \caer{} can be enabled from the first optimization step without retuning the loss scale, and its focusing signal is refreshed at every step instead of being fixed in advance. Across action-conditioned tasks with distinct control modalities, it converges to better solutions than the matched uniform-MSE baseline, with stronger visual quality and physical consistency. As training proceeds, the focus contracts onto causal interaction regions and those dynamics are predicted with progressively greater accuracy. The resulting world model therefore captures not only sharper appearance, but also when interactions occur, how objects respond to interventions, and how agents affect the surrounding world.

\section{Methodology}
\label{Sec:method}

\caer{} allocates gradient to locations where the predicted future is causally sensitive to
the action, a signal read online from the world model itself without segmentation, tracking,
optical flow, or contact labels. Motion cues identify \emph{what moves}, whereas control
requires \emph{what the action governs}: camera shake and exogenous motion are salient under
the former yet carry no action-conditional risk. Crucially, the weight redistributes a fixed
coefficient mass instead of enlarging it, which turns the comparison with uniform training
into a pure allocation question and lets us characterize, at equal mass, when the reallocation
provably wins (Fig.~\ref{fig:caer_framework}).

\subsection{Problem Formulation} \label{Sec:formulation} Let the backbone be a pretrained video generator and $z_0\in\mathbb{R}^{B\times C\times T\times h\times w}$ the video latent, where $t=0$ denotes the reference frame and $t=1,\ldots,T-1$ the supervised future. We use the linear flow-matching path \begin{equation} \begin{aligned} z_\tau &= (1-\tau)z_0+\tau\epsilon,\\ v_{\mathrm{tgt}} &= \epsilon-z_0, \qquad \epsilon\sim\mathcal{N}(0,I), \end{aligned} \end{equation} where $\tau$ follows the backbone's timestep distribution. The model predicts $v_\theta(z_\tau;c_A)$ conditioned on $c_A=\{\mathbf{o}_0,\mathbf{p},\mathcal{A}\}$, containing the reference observation, language instruction, and action condition. Uniform flow matching averages token errors, allocating gradient by token count rather than by downstream utility: static background tokens dominate the coefficient mass although generation failures concentrate on sparse actor--object interactions.

\subsection{Action Conditioning and Effect Estimation} \label{Sec:action_injection} 

\noindent\textbf{Action conditioning.} A frame-aligned action sequence $A=[a_1,\ldots,a_K]$, with $a_k\in\mathbb{R}^{d_a}$, is mapped to a backbone-compatible condition $\mathcal{A}=\phi(A)$. The encoder $\phi$ follows the backbone's control interface, either spatial, where frame-wise action maps are VAE-encoded onto the video latent grid, or numerical, where action vectors modulate each Transformer block (Table~\ref{tab:settings}). New pathways are zero-initialized where possible, preserving the pretrained image-to-video function at initialization.

\noindent\textbf{Null action.} Reweighting only requires that removing $\mathcal{A}$ produces a well-defined prediction $v_\theta(z_\tau;c_{\emptyact})$, where $c_{\emptyact}=\{\mathbf{o}_0,\mathbf{p},\emptyact\}$. We realize $\emptyact$ by zeroing or masking the injected action features, whichever form the interface exposes. Because an action-free input would otherwise be out of distribution, we independently disable action injection with probability $p_{\mathrm{drop}}=0.10$ and predict the same target from $c_{\emptyact}$. Similar to classifier-free guidance~\cite{ho2022classifierfree}, this trains a usable null branch. Here it also keeps both \caer{} queries in distribution and provides the recall floor in Eq.~\eqref{eq:recall_floor}. 

\noindent\textbf{Action effect map.} We evaluate a controlled counterfactual: under the same history and noisy latent, where does the predicted velocity change when the true action is removed? Using shared noise $\epsilon'$, we build a latent at a fixed intermediate time
$\tau_S=0.50$ by $z_{\tau_S}=(1-\tau_S)z_0+\tau_S\epsilon'$, and reduce the two responses
along the channel dimension:
\begin{equation}
S=\left\|
v_\theta(z_{\tau_S};c_A)-v_\theta(z_{\tau_S};c_{\emptyact})
\right\|_2
\label{eq:action_effect}
\end{equation}
with $S\in\mathbb{R}^{B\times1\times T\times h\times w}$. Sharing $\epsilon'$ removes sampling variation, while fixing $\tau_S$ prevents the scale and spatial structure of $S$ from drifting with the random training timestep. The two evaluations are extra forwards under no-gradient mode, run alongside the single gradient-carrying forward at the sampled $\tau$ that produces the loss. On the linear path $\hat z_0=z_{\tau_S}-\tau_S v_\theta$, so $S=\tau_S^{-1}\|\delta_\theta\|_2$
for the action-induced increment between the model's two clean-state predictions,
$\delta_\theta=\hat z_0(z_{\tau_S};c_A)-\hat z_0(z_{\tau_S};c_{\emptyact})$,
which is the model's current estimate of
\begin{equation}
\delta=\mathbb{E}\!\left[z_0\mid z_{\tau_S},c_A\right]
-\mathbb{E}\!\left[z_0\mid z_{\tau_S},c_{\emptyact}\right].
\label{eq:delta}
\end{equation}
Thus $S$ measures how strongly the model's predicted future depends on the action, which is
what separates it from flow or segmentation: a token with negligible $\delta_i$ carries little
action-conditional risk regardless of its pixel-space motion. Since $z_{\tau_S}$ is a noised
version of the realized future and hence itself downstream of the action, Eq.~\eqref{eq:delta}
is an action-conditional predictive contrast rather than an interventional quantity in the
sense of do-calculus; we use \emph{causal} in this operational sense throughout, which suffices
because the objective needs only a ranking of tokens by residual action dependence, not an
identified causal graph.

\subsection{Focused Objective and Optimization Effect} \label{Sec:objective} 
\noindent\textbf{Budget-neutral weight.} 
Let $\Omega_b=\{(t,x):t=1,\ldots,T-1\}$ be the $|\Omega_b|=(T-1)hw$ future tokens of sample $b$, and $\mu^{(b)}(F)=|\Omega_b|^{-1}\sum_{(t,x)\in\Omega_b}F^{b,t}_x$ the
future-token mean of a scalar map $F$. Because the magnitude of $S$ changes with training progress and action amplitude,
we normalize it independently for each sample:
\begin{equation}
\rho^{b,t}_x
=
\frac{S^{b,t}_x}
{\max\!\left\{\mu^{(b)}(S),\varepsilon_{0}\right\}},
\qquad
\mu^{(b)}(\rho)=1 ,
\label{eq:rho}
\end{equation}
with a small constant $\varepsilon_{0}>0$ for numerical safety. The unit-mean constraint fixes
the total coefficient mass to that of uniform training, so reweighting redistributes coefficient mass
within each sample without increasing its total mass or shifting mass across samples. The
construction of $\rho$ is detached from the backward graph.

\noindent\textbf{Recall floor.} A token with zero measured effect would receive no gradient, so a false negative could never be corrected, whereas a false positive only spends part of one update; the error is therefore asymmetric. Action dropout removes this failure mode at no additional cost. On dropped samples we skip Eq.~\eqref{eq:action_effect} and set $\rho\equiv1$, supervising every token with unit weight through the null branch, which shares all parameters with the action branch. Averaged over the dropout variable, the coefficient applied to token $(t,x)$ is
\begin{equation}
\begin{aligned}
\bar\rho^{b,t}_x &= (1-p_{\mathrm{drop}})\rho^{b,t}_x+p_{\mathrm{drop}},\\
\bar\rho^{b,t}_x &\ge p_{\mathrm{drop}}>0,
\qquad
\mu^{(b)}(\bar\rho)=1 ,
\end{aligned}
\label{eq:recall_floor}
\end{equation}
so the effective weight field is strictly positive and no token is permanently excluded from optimization, while the coefficient mass is preserved exactly rather than inflated by a hand-set lower bound; because
dropped samples are supervised under $c_{\emptyact}$, Eq.~\eqref{eq:recall_floor} is the
coefficient reaching token $(t,x)$ through the shared parameters. With $c_b$ the condition used
in sample $b$'s main forward, and $\rho\equiv1$ on dropped samples, the token error is \begin{equation} \ell^{b,t}_x = \left\| v_\theta(z_\tau;c_b)^{b,t}_x - v_{\mathrm{tgt}}{}^{b,t}_x \right\|_2^2. \end{equation} Since $\sum_{(t,x)\in\Omega_b}\rho^{b,t}_x=|\Omega_b|$, no additional normalization is required: \begin{equation} \begin{aligned} \mathcal{L}_{\mathrm{focus}} = \frac{1}{B} \sum_{b=1}^{B} \frac{1}{|\Omega_b|} \sum_{(t,x)\in\Omega_b} \operatorname{sg} \!\left(\rho^{b,t}_x\right) \ell^{b,t}_x . \end{aligned} \label{eq:focus_loss} \end{equation} 

\noindent\textbf{Allocation gain at equal coefficient mass.}
We drop the sample index and write $g_i=\nabla_\theta\ell_i$ for $i\in\Omega$. For a tolerance $\kappa\ge0$, let $R=\{i:\|\delta_i\|>\kappa\}$ collect the tokens whose predicted future is appreciably action-dependent, and $R^{c}=\Omega\setminus R$ the background.
Define the interaction risk, the region gradients, and the utility of token $i$ as
\begin{equation}
\mathcal{J}_R
=\frac{1}{|R|}\sum_{i\in R}\ell_i,
\quad
G_R=\frac{1}{|\Omega|}\sum_{i\in R}g_i,
\quad
a_i=\left\langle\nabla\mathcal{J}_R,g_i\right\rangle,
\end{equation}
with $G_{R^{c}}$ defined analogously, so that
$\nabla\mathcal{J}_R=\frac{|\Omega|}{|R|}G_R$ and
$G_{\mathrm{uni}}=G_R+G_{R^{c}}$.
Comparing the equal-mass updates
$G_{\mathrm{uni}}=|\Omega|^{-1}\sum_i g_i$ and
$G_\rho=|\Omega|^{-1}\sum_i\rho_i g_i$ to first order and using $\mu(\rho)=1$,
\begin{equation}
\begin{aligned}
&\mathcal{J}_R(\theta-\eta G_{\mathrm{uni}})
-\mathcal{J}_R(\theta-\eta G_\rho)\\
&\qquad=\eta\,\operatorname{Cov}_i(\rho_i,a_i)+O(\eta^2),
\end{aligned}
\label{eq:allocation}
\end{equation}
where the covariance is taken over tokens drawn uniformly from $\Omega$.
Positive covariance between weight and utility is therefore exactly the condition, to first order in $\eta$, 
under which focused weighting reduces interaction risk faster than uniform
weighting at identical coefficient mass.

Equation~\eqref{eq:action_effect} supplies that correlation. For $i\in R$ the utility decomposes as
\begin{equation}
a_i=\frac{1}{|R|}
\Big(
\|g_i\|_2^2
+\!\!\sum_{j\in R\setminus\{i\}}\!\!
\left\langle g_j,g_i\right\rangle
\Big),
\end{equation}
whose diagonal term is strictly positive whenever the token is not yet solved,
whereas a token in $R^{c}$ enters $\mathcal{J}_R$ only through parameter sharing and contributes cross terms of no systematic sign.
Since $S_i\rightarrow\|\delta_i\|/\tau_S$ as the estimate improves, $\rho$ places above-average weight precisely on the tokens that carry this positive diagonal term.
Only a positive association between $\rho$ and $a$ is required, not calibrated magnitudes, so the argument tolerates estimation error in $S$, and it is the association across tokens rather than the sign of any single $a_i$ that matters.

\begin{table}[htbp]
\centering
\small
\begin{tabular}{@{}
>{\raggedright\arraybackslash}p{2.8cm}
>{\raggedright\arraybackslash}p{2.5cm}
>{\raggedright\arraybackslash}p{2.4cm}
>{\raggedright\arraybackslash}p{4.5cm}
>{\raggedright\arraybackslash}p{2.6cm}
@{}}
\toprule
Task setting & Training source & Samples (clips $\times$ frames) & Action injection & Evaluation benchmark \\
\midrule
LIBERO~\cite{liu2023libero}
& LIBERO simulator rollouts & 44,166 $\times$ 17
& Low-dimensional actions $\rightarrow$ MLP modulation
& WorldArena~\cite{shang2026worldarena} \\
RoboTwin~\cite{chen2025robotwin}
& RoboTwin 2.0 & 50,016 $\times$ 17
& 14-D actions $\rightarrow$ MLP modulation
& WorldArena~\cite{shang2026worldarena} \\
Camera Control
& RealEstate10K\newline\cite{zhou2018stereo} & 30,016 $\times$ 81
& Camera trajectory $\rightarrow$ Control Adapter
& iWorld-Bench~\cite{fang2026iworldbench} \\
PoseAnything~\cite{wang2025poseanything}
& Official training set & 51,792 $\times$ 17
& Pose maps $\rightarrow$ VAE visual branch
& VBench~\cite{vbench} \\
\bottomrule
\end{tabular}
\caption{Training and evaluation settings. All experiments use Wan 2.2 5B as the backbone.}
\label{tab:settings}
\end{table}

\noindent\textbf{Escaping cancellation.}
Uniform descent halts wherever $G_{\mathrm{uni}}=G_R+G_{R^{c}}=0$, which can occur
with $G_R\neq0$: sparse interaction gradients are cancelled by background gradients even
though $\nabla\mathcal{J}_R\propto G_R\neq0$ leaves the interaction risk locally reducible.
Let $\alpha$ and $\beta$ denote the mean weights over $R$ and $R^{c}$. Unit mass forces
\begin{equation}
\frac{|R|}{|\Omega|}\alpha
+\frac{|R^{c}|}{|\Omega|}\beta=1,
\end{equation}
so a single one-sided input suffices: (A1) the effect map is informative in the weak sense that $\alpha>1$. Unit mass then forces $\beta<1$, and $\alpha>\beta$ follows from Eq.~\eqref{eq:rho} rather than from a separate assumption about the background. Treating $\rho$ as constant within each region, at such a point
$\theta_u$
\begin{equation}
\begin{aligned}
G_\rho(\theta_u)
&=\alpha G_R+\beta G_{R^{c}}
=(\alpha-\beta)G_R\neq0,\\[2pt]
\left\langle\nabla\mathcal{J}_R,G_\rho(\theta_u)\right\rangle
&=(\alpha-\beta)\frac{|\Omega|}{|R|}\|G_R\|_2^2>0 .
\end{aligned}
\label{eq:decancel}
\end{equation}
Focused weighting therefore produces a strict first-order decrease of the interaction risk
under (A1) alone, with no assumption on the alignment between $G_R$ and $G_{R^{c}}$, and the guarantee is quadratic in the very
gradient that uniform training cancelled away. Because $\rho$ is recomputed and detached at
every step, the result applies to the surrogate optimized at that step.

\noindent\textbf{Shared optimum and variance.}
After stop-gradient the weights are constants, and averaging over the dropout variable gives $\bar\rho\ge p_{\mathrm{drop}}>0$, so the population objective is a strictly positive weighted sum of token-wise Bregman divergences.
Because $\rho$ is built from $z_0$, it is not measurable with respect to the conditioning
alone, so whether reweighting moves the population optimum is a substantive question. Writing
$\mathbb{E}_c[\cdot]$ for the expectation given $(z_\tau,c)$, the per-token minimizer is
available in closed form and its deviation from the uniform optimum
$v^{*}=\mathbb{E}[v_{\mathrm{tgt}}\mid z_\tau,c]$ is exactly one covariance,
\begin{equation}
v^{\rho}_i=\frac{\mathbb{E}_c\!\left[\bar\rho_i v_{\mathrm{tgt},i}\right]}
{\mathbb{E}_c\!\left[\bar\rho_i\right]},
\qquad
v^{\rho}_i-v^{*}_i=\frac{\operatorname{Cov}_c\!\left(\bar\rho_i,v_{\mathrm{tgt},i}\right)}
{\mathbb{E}_c\!\left[\bar\rho_i\right]},
\label{eq:tilt}
\end{equation}
taken componentwise between the scalar weight and the vector target. Uniform flow matching is
recovered whenever the effective weight is uncorrelated with the aleatoric part of the target,
and any violation is damped by a denominator bounded below by $p_{\mathrm{drop}}$ through
Eq.~\eqref{eq:recall_floor}. The construction keeps the numerator small by design: both
branches share $z_{\tau_S}$, so common structure cancels and only the action-induced difference
survives; the channel norm discards its sign, removing direct coupling to a signed residual;
and the map is read at a fixed $\tau_S$ from an independent draw $\epsilon'$, so it is not a
function of the noise defining $v_{\mathrm{tgt}}$ at $\tau$. A dependence on $z_0$ persists
because $z_{\tau_S}$ contains it, and Eq.~\eqref{eq:tilt} states precisely how much that can
matter; up to this bounded tilt, the points escaped in Eq.~\eqref{eq:decancel} are
optimization-induced stationary points rather than alternative population optima. The remaining
cost is second order, entering through the second moment $\mu(\rho^{2})=1+\chi^{2}(\rho\,\|\,\mathrm{unif})$,
which bounds the variance inflation of the token-averaged gradient: it stays near one while
$\rho\approx1$ early in training and grows only as the weight concentrates. As $\theta$
improves, its counterfactual estimate sharpens and the covariance in
Eq.~\eqref{eq:allocation} can increase, while Eq.~\eqref{eq:recall_floor} keeps every token
receiving gradient, closing a self-consistent loop that needs no external annotation.

\begin{table}[htbp]
\centering
\setlength{\tabcolsep}{2.5pt}
\caption{\footnotesize\textbf{Uniform MSE versus \caer{} on iWorld-Bench~\cite{fang2026iworldbench} for camera control.} Aggregate plus generation-quality and trajectory-following metrics; memory metrics are omitted. Higher is better; best in \textbf{bold}.}
\label{tab:main_results_iworld}
\resizebox{\textwidth}{!}{%
\begin{tabular}{@{}llc cccc cc@{}}
\toprule
\multicolumn{2}{c}{} & \multicolumn{1}{c}{Aggregate} &
\multicolumn{4}{c}{Generation Quality} &
\multicolumn{2}{c}{Trajectory Following} \\
\cmidrule(lr){3-3}\cmidrule(lr){4-7}\cmidrule(lr){8-9}
& Objective & Avg. &
\makecell{Image\\Quality} & \makecell{Brightness\\Consistency} & \makecell{Color Temp.\\Constraint} & \makecell{Sharpness\\Retention} &
\makecell{Motion\\Smoothness} & \makecell{Trajectory\\Accuracy} \\
\midrule
\multirow{2}{*}{Camera Control} & Uniform MSE
& 0.6412 & \textbf{0.6898} & 0.5513 & 0.5196 & \textbf{0.4752} & 0.9902 & \textbf{0.6211} \\
& \caer{}
& \textbf{0.6614} & 0.6804 & \textbf{0.6693} & \textbf{0.6627} & 0.4150 & \textbf{0.9934} & 0.5474 \\
\bottomrule
\end{tabular}
}\vspace{0.3em}

\caption{\footnotesize\textbf{Uniform MSE versus \caer{} on WorldArena~\cite{shang2026worldarena} for LIBERO and RoboTwin.} EWMScore and all 15 constituent metrics across six dimensions. Higher is better; best within each task in \textbf{bold}.}
\label{tab:main_results_worldarena}
\resizebox{\textwidth}{!}{%
\begin{tabular}{@{}llc ccc ccc ccc cc cc cc@{}}
\toprule
\multicolumn{2}{c}{} & \multicolumn{1}{c}{Aggregate} &
\multicolumn{3}{c}{Visual Quality} &
\multicolumn{3}{c}{Motion Quality} &
\multicolumn{3}{c}{Content Consistency} &
\multicolumn{2}{c}{Physics Adherence} &
\multicolumn{2}{c}{3D Accuracy} &
\multicolumn{2}{c}{Controllability} \\
\cmidrule(lr){3-3}\cmidrule(lr){4-6}\cmidrule(lr){7-9}\cmidrule(lr){10-12}
\cmidrule(lr){13-14}\cmidrule(lr){15-16}\cmidrule(lr){17-18}
& Objective & \makecell{EWM\\Score} &
\makecell{Image\\Quality} & \makecell{Aesthetic\\Quality} & \makecell{JEPA\\Similarity} &
\makecell{Dynamic\\Degree} & \makecell{Flow\\Score} & \makecell{Motion\\Smoothness} &
\makecell{Subject\\Consistency} & \makecell{Background\\Consistency} & \makecell{Photometric\\Consistency} &
\makecell{Interaction\\Quality} & \makecell{Trajectory\\Accuracy} &
\makecell{Depth\\Accuracy} & Perspectivity &
\makecell{Instruction\\Following} & \makecell{Semantic\\Alignment} \\
\midrule
\multirow{2}{*}{LIBERO} & Uniform MSE
& 57.66
& 0.3655
& 0.4950
& 0.5472
& 0.1383
& 0.0388
& 0.4853
& 0.6050
& 0.6280
& 0.6829
& \textbf{0.5900}
& 0.8111
& \textbf{0.9805}
& \textbf{0.8080}
& 0.5520
& 0.9207 \\

& \caer{}
& \textbf{61.79}
& \textbf{0.3694}
& \textbf{0.5076}
& \textbf{0.5639}
& \textbf{0.1623}
& \textbf{0.0549}
& \textbf{0.5020}
& \textbf{0.7624}
& \textbf{0.7915}
& \textbf{0.8397}
& 0.5885
& \textbf{0.8475}
& 0.9793
& 0.8077
& \textbf{0.5615}
& \textbf{0.9300} \\
\midrule
\multirow{2}{*}{RoboTwin} & Uniform MSE
& 62.35 & 0.5200 & \textbf{0.3895} & \textbf{0.8323} & 0.4667 & 0.2738 & 0.7930 & \textbf{0.8234} & \textbf{0.8967} & \textbf{0.1232} & 0.6505 & \textbf{0.2781} & 0.8093 & 0.9071 & \textbf{0.6929} & \textbf{0.8955} \\
& \caer{}
& \textbf{63.13} & \textbf{0.5502} & 0.3879 & 0.8187 & \textbf{0.5268} & \textbf{0.3251} & \textbf{0.8231} & 0.8125 & 0.8957 & 0.0986 & \textbf{0.6620} & 0.2610 & \textbf{0.8212} & \textbf{0.9180} & 0.6860 & 0.8825 \\
\bottomrule
\end{tabular}
}\vspace{0.3em}

\caption{\footnotesize\textbf{Uniform MSE versus \caer{} on VBench~\cite{vbench} for PoseAnything.} Results on six fine-grained quality dimensions. Higher is better; best in \textbf{bold}.}
\label{tab:main_results_vbench}
\resizebox{\textwidth}{!}{%
\begin{tabular}{@{}ll ccccccc@{}}
\toprule
& Objective & Aggregate &
\makecell{Subject\\Consistency} & \makecell{Background\\Consistency} &
\makecell{Motion\\Smoothness} & \makecell{Dynamic\\Degree} &
\makecell{Aesthetic\\Quality} & \makecell{Imaging\\Quality} \\
\midrule
\multirow{2}{*}{PoseAnything} & Uniform MSE
& 0.7422 & 0.8460 & \textbf{0.9273} & \textbf{0.9686} &
0.74 & \textbf{0.4348} & 0.5366 \\
& \caer{}
& \textbf{0.7746} & \textbf{0.8828} & 0.9202 & 0.9608 &
\textbf{0.86} & 0.4269 & \textbf{0.5971} \\
\bottomrule
\end{tabular}
}
\vspace{0.3em}
\end{table}

\section{Experiments}
\label{Sec:Experiments}

We evaluate \caer{} against uniform MSE under a matched training protocol across four action-conditioned settings, then analyze key hyperparameters and the evolution of the focus map during training.


\subsection{Experimental Settings}
\label{sec:exp_settings}

\noindent\textbf{Data and control interfaces.}
All four settings use Wan 2.2 5B~\cite{wan}; their data, control interfaces, and evaluation benchmarks are summarized in Table~\ref{tab:settings}. Robot actions are temporally aligned with the video clips and injected through MLP-based DiT modulation. Camera trajectories are projected into patch-level control features, whereas frame-wise pose maps are VAE-encoded to align with the video latent grid.

\noindent\textbf{Matched training protocol.}
Within each setting, uniform MSE and \caer{} share the initialization, action pathway, training samples and order, optimizer, schedule, batch size, resolution, and training steps; only the objective changes. All experiments use eight NVIDIA H20 GPUs. LIBERO, RoboTwin, and camera control are evaluated after 5,000 optimization steps, while PoseAnything is evaluated after 1,000 steps. Unless otherwise stated, \caer{} uses $p_{\mathrm{drop}}=10\%$ and $\tau_S=0.50$. Non-dropped samples run one gradient-carrying forward at the sampled $\tau$ for the loss plus two no-gradient forwards at $\tau_S$ for the action effect, adding no backward pass and no external model; dropped samples train the null-action branch with uniform flow matching. We report each benchmark's official metrics and, where feasible, mean and standard deviation across repeated runs, wall-clock time, peak memory, and steps to convergence.

\subsection{Main Comparison Across Action-Conditioned World Model}
\label{sec:main_results}

To comprehensively assess the generality of \caer{} as a training paradigm, we compare it with uniform MSE across four heterogeneous action-conditioned tasks and their corresponding benchmarks. Each pair uses the same initialization, data, action interface, and optimization schedule, with the objective as the only difference. Tables~\ref{tab:main_results_iworld},~\ref{tab:main_results_worldarena}, and~\ref{tab:main_results_vbench} report the official metrics of iWorld-Bench, WorldArena, and VBench, covering visual and motion quality, physical consistency, controllability, and semantic alignment.

Across all four AC-WM settings, \caer{} consistently outperforms the matched uniform-MSE baseline under the same backbone, data, and training schedule. Gains appear not only in aggregate scores but also across fine-grained dimensions that stress action following, interaction fidelity, and physical consistency, indicating that action-causal reweighting systematically redirects capacity toward tokens that determine controllable dynamics rather than background residuals. Because the improvement holds for robot manipulation (LIBERO and RoboTwin), camera control, and pose-conditioned generation alike, these results support \caer{} as a more general and better-suited training objective for action-conditioned world models than space--time-uniform MSE.

\begin{figure}[htbp]
\centering
\includegraphics[width=\linewidth]{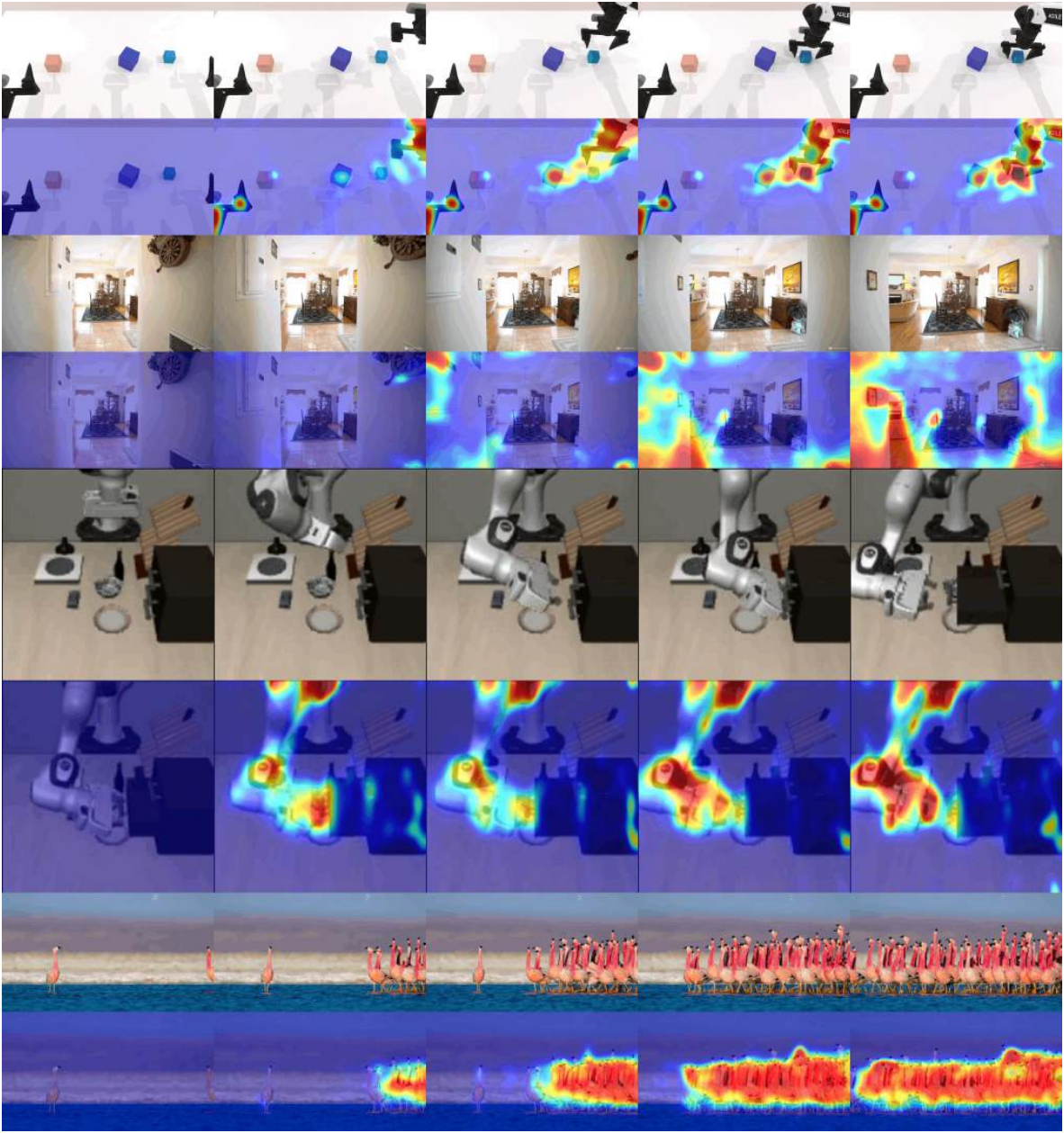}
\caption{\textbf{Qualitative Results.} Generated videos and their  \caer{} loss-supervision maps across four distinct action-conditioning modalities. Darker red regions indicate larger loss values, whereas darker purple regions indicate smaller loss values.}
\label{fig:qualitative_results}

\end{figure}

Figure~\ref{fig:qualitative_results} confirms that the loss concentrates on action-relevant
regions across control modalities while downweighting already-modeled content. Benchmark-specific
qualitative comparisons are provided in Appendix~\ref{app:qualitative_comparisons}: camera control
is shown in Figure~\ref{fig:app_camera_qual}, RoboTwin in Figure~\ref{fig:app_robotwin_qual},
PoseAnything in Figure~\ref{fig:app_poseanything_qual}, and LIBERO in
Figure~\ref{fig:app_libero_qual}. These examples provide a closer view of how \caer{} shifts
supervision toward action-relevant regions under different control interfaces. Quantitatively, \caer{} improves the score from 0.6412 to 0.6614 on camera control, from 57.66 to 61.79 on LIBERO, from 62.35 to 63.13 on RoboTwin, and from 0.7422 to 0.7746 on PoseAnything. The largest gains occur in motion- and interaction-sensitive dimensions, including dynamic degree, flow quality, content consistency, and imaging quality. Several strong appearance or trajectory metrics remain unchanged or slightly decrease. This suggests that \caer{} reallocates capacity from easy, action-independent content toward unresolved action-conditioned dynamics rather than optimizing every token uniformly, improving world-model performance.

\subsection{Training Hyperparameter Analysis}
\label{sec:hyperparameter}

As shown in Table~\ref{tab:hyperparameter}, we next analyze the sensitivity of \caer{} to action dropout and the fixed-noise operating point. For action dropout, we compare $p_{\mathrm{drop}}\in\{5\%,10\%,15\%,20\%\}$ while fixing $\tau_S=0.50$. For the effect-map operating point, we compare $\tau_S\in\{0.25,0.50,0.75\}$ while fixing $p_{\mathrm{drop}}=10\%$. Changing one factor at a time separates null-branch quality from noise-level sensitivity.

The results show a clear trade-off in the action dropout rate. A very small dropout rate insufficiently trains the null-action branch, whereas an excessively large rate causes action-irrelevant regions to dominate the reweighting signal. Empirically, approximately \(10\%\) provides a reasonable balance.

The results also highlight the importance of selecting an appropriate noise level \(\tau_S\). Extremely small or large values either suppress the difference between the two branches or weaken scene-specific localization. In comparison, \(\tau_S = 0.5\) preserves both action dependence and scene context, producing the most informative action-effect map.

\begin{table}[t]
\centering
\small
\begin{tabular}{llcc}
\toprule
Factor & Value & WorldArena $\uparrow$ & iWorldBench $\uparrow$ \\
\midrule
\multirow{4}{*}{Action dropout} & $5\%$ & 62.83 & 0.5447 \\
 & $\mathbf{10\%}$ &  $\mathbf{63.13}$ & $\mathbf{0.6614}$ \\
 & $15\%$ & 61.91 & 0.5797 \\
 & $20\%$ & 61.40 & 0.5470 \\
\midrule
\multirow{3}{*}{Fixed time $\tau_S$} & $0.25$ & 61.09 & 0.5652 \\
 & $\mathbf{0.50}$ & $\mathbf{63.13}$ & $\mathbf{0.6614}$ \\
 & $0.75$ & 62.27 & 0.6167 \\
\bottomrule
\end{tabular}
\caption{Sensitivity of \caer{} to action dropout and fixed noise $\tau_S$ on RoboTwin (WorldArena) and camera control (iWorld-Bench). Bold entries denote the default configuration.}
\label{tab:hyperparameter}

\end{table}

\subsection{Toy Study: Score Evolution During Training}
\label{sec:toy_study}

\begin{figure}[htbp]
\centering
\includegraphics[width=\linewidth]{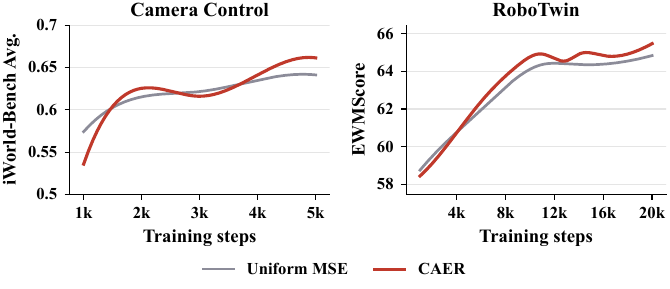}
\caption{\textbf{Score evolution across intermediate training checkpoints.}
iWorld-Bench on camera control and WorldArena on RoboTwin are shown under
\caer{} and uniform MSE. Uniform MSE often leads at the earliest checkpoints,
whereas \caer{} catches up and becomes higher later, while retaining small
local fluctuations throughout training.}
\label{fig:toy_progression}

\end{figure}

Finally, we evaluate intermediate checkpoints throughout training on both benchmarks to
characterize how the two objectives trade off over optimization. Figure~\ref{fig:toy_progression}
plots iWorld-Bench on camera control and WorldArena on RoboTwin, with \caer{} and uniform MSE
sharing the same initialization, data order, and schedule. The four curves exhibit an
early-lead-to-late-lead crossover pattern. At the earliest checkpoints, uniform MSE often scores
higher, since spreading the coefficient mass over all tokens quickly reduces abundant background residuals
that appearance-driven metrics reward. \caer{} instead allocates part of that mass to the harder
interaction tokens. As the action-conditional dynamics become better learned, \caer{} catches up
and generally overtakes uniform MSE on both benchmarks. Although the curves retain local
fluctuations, \caer{} maintains a positive late-stage margin, while uniform MSE largely plateaus
with interaction errors remaining unresolved. This crossover is consistent with the behavior
predicted by the self-improving loop: a sharper $\theta$ yields a sharper effect map and directs
more of the fixed coefficient mass toward what remains unresolved.

\section{Related Work}
\label{related_work}

\noindent\textbf{Video-generative world models.} Large video diffusion and flow-matching models, including Wan~\cite{wan}, HunyuanVideo~\cite{hunyuanvideo}, CogVideoX~\cite{yang2024cogvideox}, and Cosmos~\cite{cosmos}, provide the visual and temporal priors on which many recent world models are built. Action-conditioned systems span camera trajectories, navigation commands, game controls, robot actions, and human motion. Genie~\cite{bruce2024genie}, Matrix-Game~\cite{he2025matrix,zhang2025matrix,wang2026matrix}, HY-World~\cite{sun2025worldplay,hy2026hy}, Cosmos-Predict~\cite{cosmospredict,ali2025world}, and Worldscape-MoE~\cite{fang2026worldscapemoe} study interactive scene exploration, while DreamGen~\cite{dreamgen}, RoboDreamer~\cite{robodreamer}, Hand2World~\cite{wang2026hand2world}, Generated Reality~\cite{xie2026generated}, WorldVLN~\cite{zhao2026worldvln}, and WorldScape Policy~\cite{su2026worldscapepolicy} extend video world models to robot, hand, body, and navigation control. Despite their different interfaces, these methods generally retain a space--time-uniform denoising objective, allowing sparse action consequences to be overwhelmed by background tokens. \caer{} addresses this shared optimization problem by redistributing supervision after action injection rather than introducing another control interface.

\noindent\textbf{External spatial supervision for interaction-aware reweighting.} External models such as SAM~\cite{kirillov2023sam} and RAFT~\cite{teed2020raft} provide segmentation and motion cues for localizing dynamic regions. MotiF~\cite{wang2025motif} reweights video objectives with optical-flow heatmaps; BridgeV2W~\cite{chen2026bridgev2w}, Mask World Model~\cite{lou2026mwm}, and Mask2Real-WM~\cite{feingold2026mask2realwm} use rendered or semantic masks; and ChronoDreamer~\cite{zhou2025chronodreamer} learns from simulator contact maps. Although effective, these pipelines require per-frame preprocessing, rendering, or simulation, and their visual proxies need not coincide with action causality. \caer{} instead extracts interaction-aware weights online from the model's own action response without external spatial annotations.

\section{Conclusion and Future Work}
We propose \caer{}, a general world-model training objective that identifies action-causal regions by
contrasting predictions with and without actions, then emphasizes unresolved interactions while
preserving the overall coefficient mass. Across four heterogeneous action-conditioned settings,
\caer{} consistently outperforms uniform MSE and develops increasingly accurate interaction focus as
training progresses. Building on this framework, future work will investigate broader mechanisms and
effects of action causality in world models, including how interventions shape learned dynamics,
generalization, and controllability, how such reweighting scales to longer horizons and larger
backbones, and how this contrast can supply signals for evaluation and data selection.

\bibliography{references}

\newpage
\appendix
\section{Additional Qualitative Comparisons}
\label{app:qualitative_comparisons}
This appendix provides benchmark-specific qualitative comparisons between uniform MSE and
\caer{} across the four action-conditioned settings. Each figure presents representative
conditioning or original frames together with the corresponding supervision maps produced by
uniform MSE and \caer{}. The same color convention as in Figure~\ref{fig:qualitative_results}
is used throughout: warmer colors indicate larger loss responses, whereas cooler colors indicate
smaller responses. Overall, \caer{} produces more spatially concentrated responses around
action-relevant regions, while uniform MSE distributes supervision more broadly across the scene.

\begin{figure}[htbp]
\centering
\includegraphics[width=\linewidth]{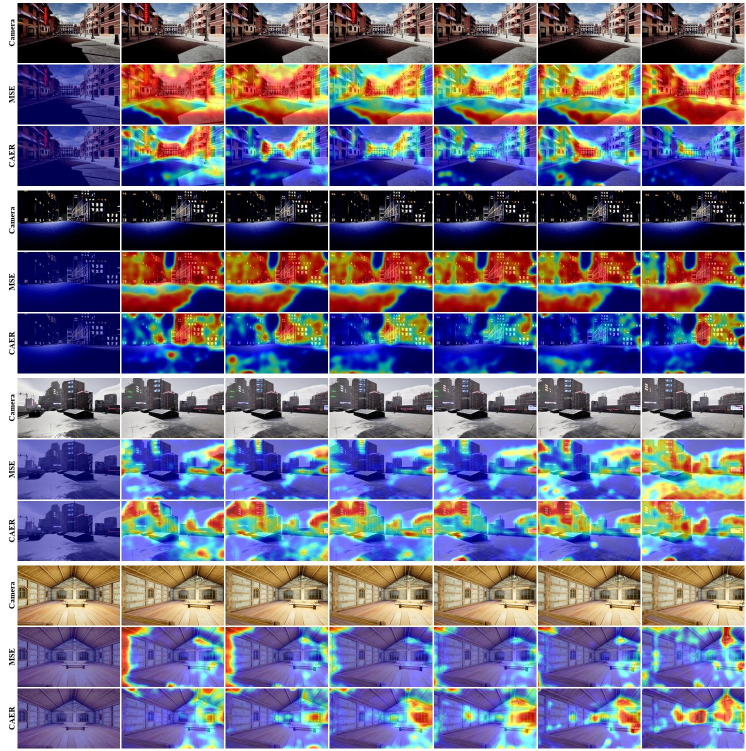}
\caption{\textbf{Camera-control qualitative comparison.}
Representative camera-control examples and their supervision maps under uniform MSE and \caer{}.
Compared with the spatially broader responses of uniform MSE, \caer{} places more emphasis on
regions associated with camera-induced motion and scene changes while suppressing already modeled
background content.}
\label{fig:app_camera_qual}
\end{figure}

\begin{figure}[htbp]
\centering
\includegraphics[width=\linewidth]{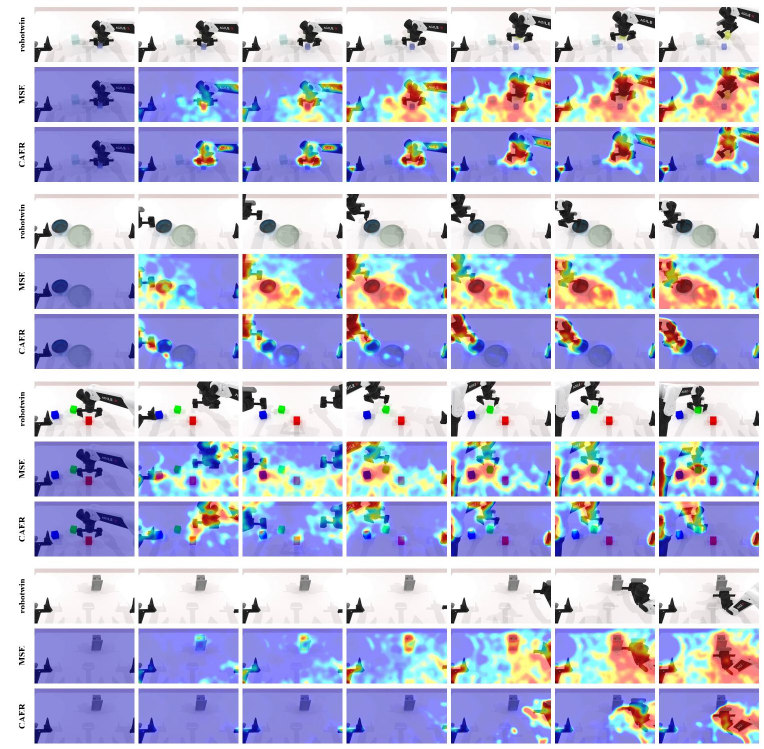}
\caption{\textbf{RoboTwin arm-manipulation qualitative comparison.}
Representative RoboTwin manipulation examples and the corresponding supervision maps under
uniform MSE and \caer{}. \caer{} more consistently concentrates responses around the robot
end-effector, manipulated objects, and interaction or contact regions, while uniform MSE assigns
substantial responses to broader scene regions.}
\label{fig:app_robotwin_qual}
\end{figure}

\begin{figure}[htbp]
\centering
\includegraphics[width=\linewidth]{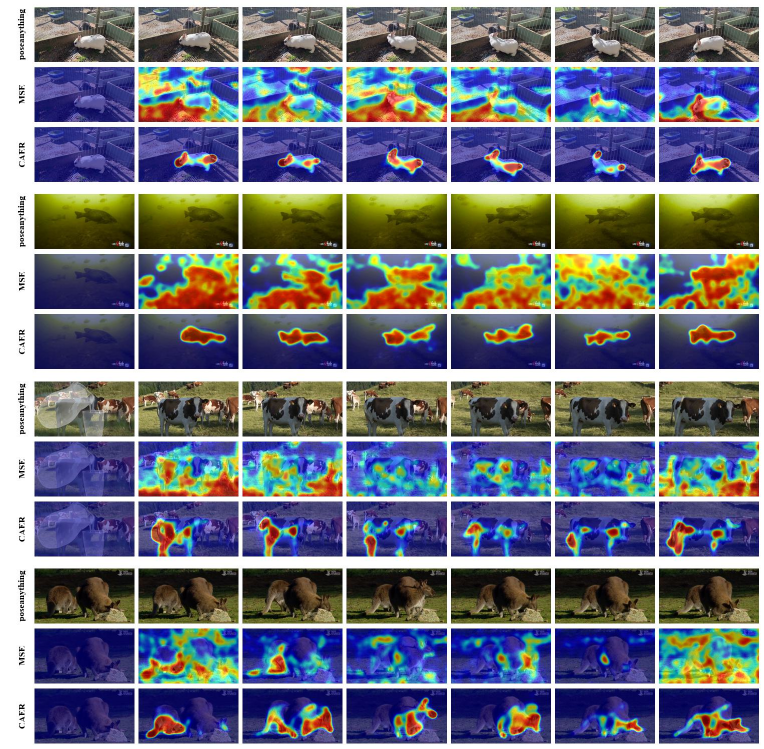}
\caption{\textbf{PoseAnything qualitative comparison.}
Representative pose-conditioned generation examples and their supervision maps under uniform MSE
and \caer{}. The \caer{} maps are more focused on articulated subjects and pose-bearing regions,
whereas uniform MSE exhibits more diffuse responses over the surrounding visual content.}
\label{fig:app_poseanything_qual}
\end{figure}

\begin{figure}[htbp]
\centering
\includegraphics[width=\linewidth]{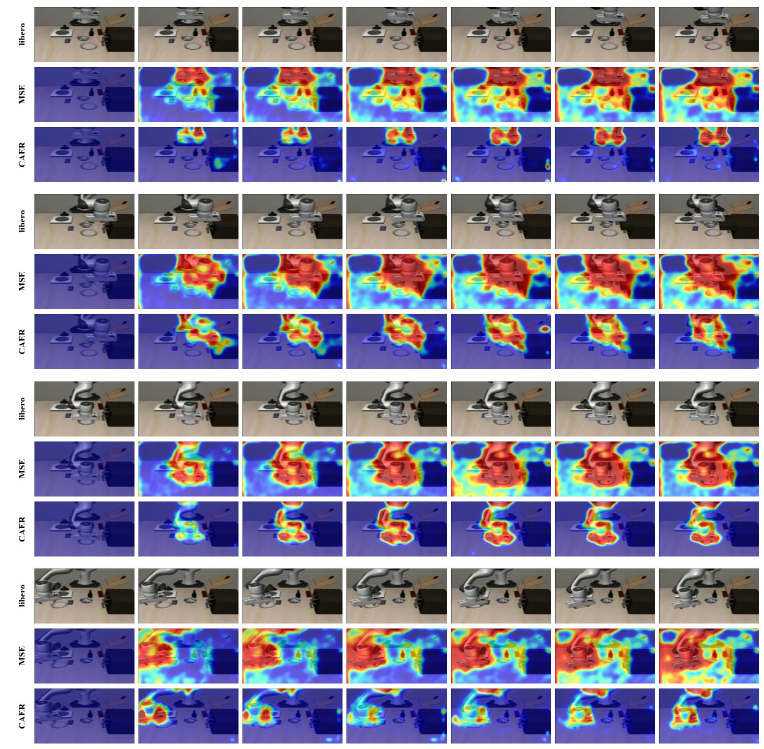}
\caption{\textbf{LIBERO manipulation qualitative comparison.}
Representative LIBERO robot-manipulation examples and the corresponding supervision maps.
\caer{} emphasizes the robot, manipulated object, and relevant workspace involved in the action,
while uniform MSE spreads supervision more broadly across the scene.}
\label{fig:app_libero_qual}
\end{figure}

\end{document}